# Identify Apple Leaf Diseases Using Deep Learning Algorithm

**Zhang Daping, Yang Hongyu, Cao Jiayu**

## Abstract:

Agriculture is an essential industry in the both society and economy of a country. However, the pests and diseases cause a great amount of reduction in agricultural production while there is no sufficient guidance for farmers to avoid this disaster. To address this problem, we apply CNNs to plant disease recognition by building a classification model.

Within the dataset of 3,642 images of apple leaves [1], We use a pre-trained image classification model Restnet34 based on Convolutional neural network (CNNs) with the Fastai framework in order to save the training time. Overall, the accuracy of classification is 93.765%.

# I. Introduction

Agriculture is one of the most important industries of one country since it plays an essential role in the production of raw materials for other industries. The usage of agricultural products can be in the fields of food, textiles, chemical industry, and so on.

However, plant disease detection is still a challenging task for farmers. Each farmer has to be able to identify diseases in plants correctly to avoid the potential risk of plant diseases. The farmers can only contact either their peasants or the assistance line for consultation. Also, it is expensive for a team of experts to detect such plant illnesses. The Inability to cure a disease due to lack of knowledge may become a main factor that caused the transferring of plant diseases because they can't be detected at an early stage. To deal with these problems existing in the agricultural field, researchers have made great progress in the diagnosis of plant leaves diseases and pests.

In this paper, we use a collected dataset from the "The Plant Pathology Challenge 2020 data set to classify foliar disease of apples" [1], which consists of 3642 images of apple leaves including healthy leaves, rust leaves, grab leaves, and so on. We divided the images into training and testing groups, then we proposed a CNNs-based model: resnet34 to trained the prediction system. We didn't follow the past works which need to train the model start from the beginning with a huge amount of data. To save the training time and remain the accuracy higher, we use the Fastai framework for actual implementation. As a result, we get an accuracy of 93.765% for the identification of apple healthy condition.

The following paper is organized into four parts. In section 2, we do a literature review on computer vision of leaf diseases identification. In section3, a detailed approach that we used to train the CNNs-based model is shown. In section 4, there is a processing of the experiment. In section 5, we conclude our system and result, then propose the future development planning.

## II. Related Work

Convolutional Neural Networks (CNNs) is the most popular architecture to deal with the computer vision problems like image classification in recent years. Since it is better than any other traditional recognition approaches, like Genetic Algorithm(GA), k-Nearest Neighbour Classifier (KNN), and Probabilistic Neural Network(PNN), to do the image classification, CNNs has been widely used in the field of plant diseases classification and identification. There is the review of literature in the following paragraphs

a) Literature summary

Wan-jie Liang, Hong Zhang et.al combined the machining learning and recognition approaches to do the "Rice Blast Disease Recognition" [2] with the CNNs architecture. They collected 5,808 samples, 2906 of them are positive while 2902 of them are negative, for CNN model training and testing. In this paper, the authors showed how CNNs performed better than local binary histogram (LBPH) and Haar-WT (Wavelet Transform), and it more paired with Softmax and SVM, larger area under curve (AUC), and better receiver operating characteristic (ROC) curves.

UDAY PRATAP SINGH et.al had made an innovation in this field. 1,070 images of mango leaves including both the healthy and the infected were collected to build a classification system of the "Mango leaves infected by the Anthracnose fungal disease" [3] . They proposed a new concept of multilayer convolutional neural network(MCNN), which contains both multiple more feed-forward layers and pooling layers. In this model, they got an accuracy of 97.13%, which is higher than other advanced technologies.

To save the training and testing time of a CNN model, researchers pay more attention to the transfer learning models, a pre-trained model with a large amount of dataset. Liu Bi et. al [4] had collected 107,366 grape leaf images from the field and the public. They proposed a novel recognition approach named DICNN, which combined the transfer learning models of InceptionV3, VGG16, and VGG19 with classifiers such as KNN, Neural Networks, Logistic regression, and SVM. Finally, they realize an accuracy of 97.22% which increases by 2.97% and 2.55% respectively compared to GoogLeNet and ResNet-34.

b) Comparison

From the previous experience, we found that many of the past works are facing the problem that it takes too much time to train a CNNs model. The main reasons may be that the researchers have to train the model start from the beginning, and even the application of the transfer learning models faces this difficulty because their frameworks like Keras and Tensorflow can't get a high computation speed sometimes. To deal with these problems, we use the Fastai framework to get higher accuracy with less training time.

## III. Methodology

### Dataset used

Appropriate data is necessary through the whole process of research. The data set of apple leaves images used in this study is obtained from The Plant Pathology Challenge 2020 data set to classify foliar disease of apples [5]. 3642 high-quality and real-life images were captured during the 2019 growing season by Cornell AgriTech, from commercially produced varieties in an unsprayed apple orchard (Geneva, New York, USA). Most of the images were of apple scab, cedar apple rust, Alternaria leaf spot, frogeye leaf spot, and healthy leaves.

The data set is complicated due to different factors including different disease categories, one-homogenous image background, different times of day, etc. Figure 2 shows the images of disease symptoms on apple leaves taken under different light conditions. Moreover, the data includes the annotated disease data set by manually annotating images of cedar apple rust (Fig.3A), apple scab (Fig. 3B), and healthy leaves (Fig. 3D). An expert plant pathologist conformed the annotations, particularly for those photos which were difficult to identify. The data set was randomly divided into training (80%) and stratified test sets (20%), with all four disease categories represented in both data sets.

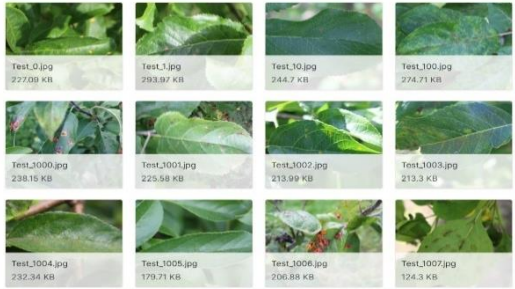

Figure 1. Part of images in the data set

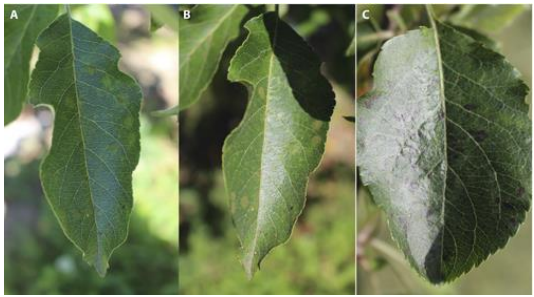

Figure 2. Images of disease symptoms on apple leave captured under different light conditions: indirect sunlight on the leaf (A), direct sunlight on the leaf (B), and strong reflection on the leaf (C).

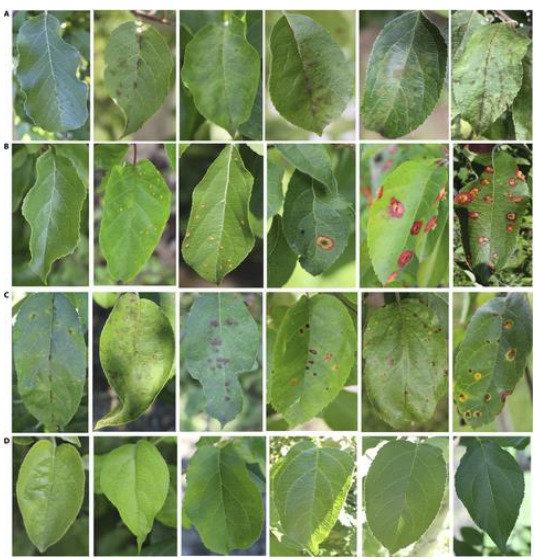

Figure 3. Sample images from the data set showing symptoms of cedar-apple rust (A), apple scab (B), multiple diseases on a single leaf (C), and healthy leaves (D).

## CNN

Convolutional neural networks (CNNs) is a class of artificial neural network that can be used for image classification. Compared to traditional algorithms such as K-nearest neighbours, Support vector machine etc., neural networks provide better functionalities for accomplishing the same tasks with higher efficiency [6]. CNNs can identify complex patterns in any given image such as lines, gradients, circles, etc. A lot of modifications can be made to any CNN model using different augmentations and using different activation functions such as Softmax and ReLU.

As a result of this transition towards deep learning for plant leaf image classification, there has been a surge in the use of popular transfer learning models such as AlexNet, GoogleNet, ResNet, Inception, DenseNet, etc. These models are pre-trained on large datasets consisting of thousands of images with multitude of classes [7]. Because of that, the transfer learning models give a head start to any modelling task on a new data set with their pre-trained weights. Practitioners always have the luxury of fine-tuning these models based on their own data sets' needs.

This paper uses ResNet architecture because of its Networks with large number (even thousands) of layers can be trained easily without increasing the training error percentage. Also, ResNets help in tackling the vanishing gradient problem using identity mapping [8].

**Evaluation**

To calculate the accuracy of our model, we use the receiver operating characteristic (ROC) curves and the area under curve (AUC) to do the evaluation. Based on the Confusion matrix, ROC consists of False Positive Rate(The probability of predicting a sample as positive incorrectly) as the horizontal axis and True Positive Rate(The probability of predicting a sample as positive correctly) as the vertical axis. If the ROC curve is above the diagonal, it means the prediction is better than the random classifier. To obtain an intuitive indicator to reflect the accuracy, we calculate the number of AUC.

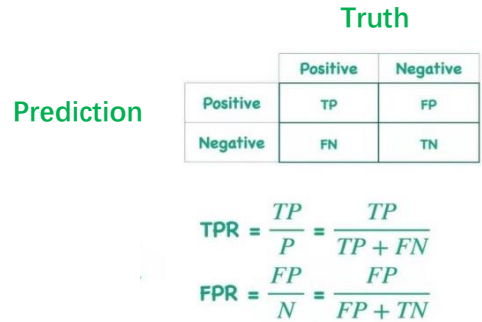

Figure 4. Calculation method of TPR and FPR

## IV. Conclusion

By combining the model of ResNet 34 and Fastai framework, we develop a system to help identify apple leaf diseases including rust, scab and multiple diseases. This algorithm will help farmers find and treat leaf diseases at the early stage in order to reduce the economic loss caused by those diseases.

For future work, we can attempt other models and optimization methods to improve the performance of the algorithm. Also, it will be useful if we can detect more kinds of diseases by using larger datasets with more disease variable. The system could also be packed to website or APP so that it could directly be used in the industry.